\documentclass[runningheads]{llncs}
\usepackage{hyperref}
\usepackage{microtype}
\usepackage{url}
\usepackage[pdftex]{graphicx}
\usepackage{siunitx}
\usepackage{amsmath,amsfonts}
\DeclareMathOperator*{\argmin}{arg\,min}

\usepackage{hyperref}
\let\svthefootnote\thefootnote
\newcommand\freefootnote[1]{%
  \let\thefootnote\relax%
  \footnotetext{#1}%
  \let\thefootnote\svthefootnote%
}

\begin{document}
%
%\title{Precision and Recall Reject Curves \\for Classification}
\title{Precision and Recall Reject Curves}
\author{Lydia Fischer\inst{1}\and
Patricia Wollstadt\inst{1}}
\authorrunning{L. Fischer and P. Wollstadt}
\institute{Honda Research Institute Europe \\
Carl-Legien-Str. 30,
63073 Offenbach am Main, Germany\\
%\url{http://www.honda-ri.de}
}
\maketitle              % typeset the header of the contribution
\begin{abstract}
For some classification scenarios, it is desirable to use only those \mbox{classification} instances a trained model associates with a high certainty.
To evaluate such high-certainty instances, previous work proposed accuracy-reject curves (ARCs).
\mbox{Reject} curves evaluate and compare the performance of different certainty \mbox{measures} over a range of thresholds for accepting or rejecting classifications.
However, in some scenarios, e.\,g., data with imbalanced class distributions, the accuracy may not be a suitable evaluation metric and measures like precision or recall may be preferable.
We therefore propose reject curves that evaluate precision and recall, the recall-reject curve and the precision-reject curve. Using prototype-based classifiers, we first validate the proposed curves on artificial benchmark data against the ARC as a baseline. We then show on benchmarks and medical, real-world data with class imbalances that the proposed precision- and recall-curves yield more accurate insights into classifier performance than ARCs.

\end{abstract}
\freefootnote{\footnotesize Acknowledgements:
The authors thank Dylan Fritts and Bradley Brown of HDMA for inspiring discussions.
The autors thank Wiebke Arlt for providing the adrenal data set and Stephan Hasler for valuable comments on the manuscript.}

\section{Introduction}
Machine learning models are used across a wide range of applications, where a common task is to train models for classification of objects.
Many of these applications are safety-critical and the reliability and trustworthiness of classifications are particularly important. Such applications come, for example, from the medical domain or high-stakes economic scenarios like logistics for just-in-time production \cite{baryannis2019supply}. %\cite{baryannis2019supply,Brintrup2019}.
One method to improve the reliability of classifiers is to calculate a certainty measure for each prediction and use only those predictions with a sufficiently high certainty.
The combination of a certainty measure and a threshold that defines the degree of certainty considered sufficient, is called \emph{reject option} \cite{chow1970optimum}.
Classifiers with reject options allow to reject data points with a unreliable classification and can improve trust in the application  \cite{DBLP:conf/ijcci/ArteltBVH22,sendhoff2020cooperative,wang2023explainable}.

When using reject options, classification performance differs for various classifiers, certainty measures, and thresholds.
To evaluate reject options, \emph{accuracy reject curves} (ARC)\cite{nadeem2009accuracy} were introduced. ARCs display the accuracy as a function of the rejection rate for a given certainty measure and classifier.
%To evaluate reject options, the authors of \cite{nadeem2009accuracy} introduced \emph{accuracy reject curves} (ARC) that display the accuracy as a function of the rejection rate for a given certainty measure and classifier.
They are a powerful tool for comparing reject options, however, in some scenarios---instead of using the accuracy to judge classifier performance---precision and recall are preferable. This is especially the case for imbalanced data sets, where the evaluation of a reject option using ARCs is inappropriate.
We introduce the precision reject curve (PRC) and the recall reject curve (RRC) to close this gap.
%To close this gap, we introduce reject curves for these alternative evaluation metrics, the precision reject curve (PRC) and the recall reject curve (RRC).

%The structure of the paper is as follows.
In the remainder of this article, we first review related work (Section~\ref{sec:rel_work}) and prototype-based classification, used in the experiments, (Section~\ref{sec:prototypes}).
Section \ref{sec:glob_rejection} introduces the framework of reject options, followed by the introduction of the PRC and RRC for evaluating reject options (Section~\ref{sec:eval}).
We demonstrate the value of the PRC and RRC in applications to artificial data with available ground-truth distribution, traditional benchmark data, and real-world medical data in Section~\ref{sec:experiments}, followed by the Conclusion in Section~\ref{sec:conclusion}.

\section{Related Work}\label{sec:rel_work}

Chow first introduced reject options with an optimal error-reject trade-off if class probabilities are known \cite{chow1970optimum}. In newer work, classification with rejection is also known as, for example, \textit{selective classification} \cite{el2010foundations}, \textit{abstention} \cite{pazzani1994trading,pietraszek2005optimizing}, and \textit{three way classification} \cite{yao2009three}.
To evaluate classifiers with reject options, the authors of \cite{nadeem2009accuracy} introduced ARCs, which are widely used today. Alternative approaches to evaluate classifiers with reject options were proposed and investigated in \cite{condessa2017performance,hanczar2019performance}.
The authors of \cite{hanczar2019performance} propose to evaluate reject options by visualising different aspects to find a suitable trade-off between either error rate and rejection rate, cost and rejection rate, or true and false positives (receiver-operator trade-off). % The authors conclude that the latter one is less convenient for evaluation method compared to the other two.  %from three perspectives: error-reject trade-off, cost-reject trade-off, and receiver-operator trade-off.
For a recent review of reject options, see \cite{reject_survey}, for a formal treatment, see \cite{franc2023optimal}.

Reject options can either be applied as a post-processing step in classification \cite{fischer2015efficient,fischer2016optimal} or can be integrated into the classifier itself, where the former offers more flexibility as it can be applied to any classifier that allows to define a certainty measure.
An instance for classification with integrated rejection is, for example, proposed by \cite{BakhtiariV22} and \cite{villmann2016self}, termed \emph{classification by component network} \cite{SaralajewHRAV19}.

Using one reject option across the whole input space is called a \emph{global reject option}. A \emph{local reject option} is defined by setting one rejection threshold per class or for different partitions of the input space \cite{fischer2016optimal}. The advantage of local reject options is that users can tune thresholds for specific classes or regions, to increase the reliability for important classes or regions. Local reject options and an efficient algorithm for finding optimal local thresholds can be found in \cite{fischer2016optimal}.
% Global reject options with certainty measures suitable for prototype-based classification were introduced in \cite{FischerHW14,fischer2015efficient}. To obtain certainty measures for prototype-based classifiers, probabilistic and deterministic approaches exist, where each approach shows advantages for certain data types \cite{FischerNVHW14}. % PW: ich verschiebe die zwei Sätze zu den Classifiern, hier versteht man eventuell noch nicht, warum  wir uns jetzt genau diesen Typ von Classifiern anschauen
% The authors of \cite{pillai2011classification} propose a generalisation for multi-label settings, which uses a $F1$-score instead of the accuracy as evaluation measure.

So far we considered reject options for offline, batch-trainable classifiers. For online learning scenarios with drift, there are only first attempts \cite{gopfert2018mitigating} which show that a reject option with a fixed threshold does not increase the performance significantly. Hence more sophisticated methods and measures are needed.

%Since it is not only important to reject unreliable decisions of a classifier, it may be relevant why an input got rejected for classification. The authors of \cite{DBLP:conf/esann/ArteltVH22,DBLP:conf/ijcci/ArteltBVH22} propose first attempts to provide an explanation. %, where the latter work uses counterfactual explanations \cite{molnar2022}.

Last, it may be relevant to understand why instances got rejected from classifications. First attempts to use explainability methods in classification with rejection are proposed by \cite{DBLP:conf/ijcci/ArteltBVH22,DBLP:conf/esann/ArteltVH22}.

% For a recent review of reject options, see \cite{reject_survey}, and for a formal treatment, see \cite{franc2023optimal}.

\section{Prototype-based Classification} \label{sec:prototypes}

To demonstrate the application of PRC and RRC, we use prototype-based classifiers in the following experiments. We consider this class of models, since it has shown good performance on the considered example data, and well-established certainty measures and global reject options are available \cite{FischerHW14,fischer2015efficient}. % Further, global reject options with certainty measures have been proposed for prototype-based classifiers \cite{FischerHW14,fischer2015efficient}, using probabilistic and deterministic approaches that each show advantages for certain data types \cite{FischerNVHW14}.

\textbf{Overview of Prototype-based Classifiers}:
We assume classification tasks in $\mathbb{R}^n$ with $Z$ classes, enumerated as $\{1, \ldots , Z\}$. Prototype-based classifiers are defined as set $W$ of prototypes $(w_j , c(w_j)) \in \mathbb{R}^n \times \{1, \ldots , Z\},\text{ and } j  \in \{1, \ldots , J \}$ that are trained on example data $X$ to represent the data and its class borders.

Every prototype $w_j$ belongs to exactly one class with its class label $c(w_j) \in \{1,\ldots Z\}$.
To classify a new data point $x$, the winner-takes-all-scheme is applied:

\begin{equation*}
    c(x) = c(w_l) \text{ with } w_l = \argmin_{w_j \in W}
d(w_j , x),
\end{equation*}
where $d$ is a distance measure, often the squared Euclidean distance.
Any prototype-based model partitions the feature space into Voronoi cells with one responsible prototype per cell.
A data point~$x$ falling into a Voronoi cell is assigned the label of the related (closest) prototype, i.\,e.\ the winner prototype.
The number of prototypes representing a class can be predefined for prototype-based models which leads to a sparse and interpretable representation of the given data~$X$.
Heuristics and cost function-based approaches are used as training techniques.

Here, we used extensions of the basic learning vector quantization (LVQ)~\cite{ritter1989self} which uses a heuristic Hebbian learning paradigm.
These extensions are the generalized matrix LVQ (GMLVQ) and local generalized matrix LVQ (LGMLVQ), and robust soft LVQ (RSLVQ), which we describe in detail in the following.

\textbf{GMLVQ and LGMLVQ:}
By formulating and optimizing explicit cost functions, extensions of LVQ are derived, namely, generalized LVQ (GLVQ) \cite{sato1995generalized} and RSLVQ \cite{seo2003soft} (next section).
For these models convergence guarantees can be given that follow directly from their derivation.
For GMLVQ \cite{biehl2007dynamics}, the distance metric is replaced by a general quadratic form, which is also learned during model training. The trained form represents a mapping that puts emphasis on the most discriminative input features and allows to reduce the feature set to the most relevant features only.
The LGMLVQ \cite{schneider2009adaptive} adds a local metric to every prototype and has shown to outperform the GMLVQ in some scenarios.

%\cite{sato1995generalized} proposed the GLVQ which was later extended to the GMLVQ and LGMLVQ.
The GLVQ is based on the formalization as minimization of the cost function
\begin{equation}
    E = \sum_{i}\Phi \left( \frac{d^+(x_i) - d^-(x_i)}{d^+(x_i) + d^-(x_i)}\right),
    \label{eq:cost_fun}
\end{equation}
%The resulting model is dubbed generalised LVQ (GLVQ).
with $\Phi$ as a monotonically increasing function, and $d^+$ and $d^-$ are the distances to the closest prototype, $w^+$ and $w^-$, of the  correct or incorrect class, for a data point $x_i$.
GLVQ optimizes the location of prototypes by means of a stochastic gradient descent based on the cost function (Eq.~\ref{eq:cost_fun}).
For a proof of the learning algorithm's validity at the boundaries of Voronoi cells see \cite{hammer2005supervised}.

The GMLVQ generalizes the GLVQ to an algorithm with metric adaptation \cite{schneider2009adaptive}.
This generalization takes into account a positive semi-definite matrix $\Lambda$ in the general quadratic form which replaces the metric $d$ of the GLVQ, i.\,e.\
$d(w_j , x) = (x - w_j )^T \Lambda(x - w_j )$.
The local version, the LGMLVQ, uses a single metric $d_j (w_j , x) = (x - w_j )^T \Lambda_j (x - w_j )$ for each prototype $w_j$.

\textbf{RSLVQ:}
RSLVQ \cite{seo2003soft} assumes that data can be modeled via a Gaussian \mbox{mixture} model with labelled types. Based on this assumption, training is performed as an optimization of the data's log-likelihood,
\begin{equation*}
    E = \sum_{i} \log p(y_i|x_i, W) = \sum_{i} \log \frac{p(x_i, y_i|W)}{p(x_i|W)},
\end{equation*}
where $p(x_i|W) = \sum_j p(w_j)\cdot p(x_i|w_j)$ is a mixture of Gaussians with uniform prior probability $p(w_j)$ and Gaussian probability $p(x_i|w_j )$ centered in $w_j$ which is isotropic with fixed variance and equal for all prototypes or, more generally, a general (possibly adaptive) covariance matrix.
The probability $p(x_i, y_i|W ) = \sum_j \delta^{c(w_j )}_{c(x_i)} p(w_j ) \cdot p(x_i|w_j )$ ($\delta^j_i$ is the Kronecker delta) describes the probability of a training sample under the current prototype distribution.
For a given prediction $\hat{y}$, RSLVQ provides an explicit certainty value $p(\hat{y}|x, W)$, due to the used probability model at the price of a higher computational training complexity.

\section{Global Reject Option}\label{sec:glob_rejection}

A reject option for a classifier is defined by a certainty measure $r$ and a threshold~$\theta$, which allows for individual samples to be rejected from classification if the classifier can not make a prediction with a certainty above the threshold.
The reject option is further called global if the threshold is constant across the whole input space, i.\,e., across all classes.
(Extending the reject options proposed in the present work to local thresholds is conceivable but beyond the scope of this article (see \cite{fischer2016optimal,kummert2016local}).)
Given a certainty measure
\begin{equation*}
    r: \mathbb{R}^n \to {\mathbb{R}}, x \longmapsto r(x) \in [0,1]
\end{equation*}
for a data point $x$ and a threshold $\theta \in \mathbb{R}$, a reject option is defined as a rejection of $x$ from classification
iff
\begin{equation*}
    r(x) < \theta .
\end{equation*}

A rejected data point will not be assigned with a predicted class label.
We denote $X_{\theta}$ as all accepted data points with a certainty higher or equal than $\theta$.
%All remaining, accepted data points with a certainty higher or equal than $\theta$, we denote by $X_{\theta}$.

In our experiments we use the certainty measures Conf (\ref{eq:conf}) and RelSim (\ref{eq:relsim}) that were proposed for prototype-based models in \cite{FischerHW14}.
Additionally, for the artificial data we consider a Bayes classifier that provides ground-truth class probabilities and serves as a baseline (see below).

\textbf{Conf:} Classifiers based on probabilistic models such as RSLVQ provide a direct certainty value of the classification with the estimated probability $\hat{p}(\cdot)$.
    \begin{equation}
        r_{\text{Conf}}(x) = \max_{1\leq j \leq Z} \hat{p}(j|x) \in (0,1]
        \label{eq:conf}
    \end{equation}

\textbf{RelSim:} The relative similarity (RelSim) \cite{FischerHW14} is based on the GLVQ cost function (\ref{eq:cost_fun}) and considers the distance of the closest prototype (the winner) $d^+$ and the distance of a closest prototype of any different class $d^-$ for a new unlabelled data point.
The winner prototype defines the class label of this new data point, if it is accepted.
The measure calculates values according to:
\begin{equation}
 r_{\text{RelSim}}(x) = \frac{d^- - d^+}{d^- + d^+} \in [0,1].
 \label{eq:relsim}
\end{equation}
Values close to one indicate a certain classification and values near zero point to uncertain class labels.
The values of $d^{\pm}$ are already calculated by the used algorithm such that no additional computational costs are caused.
Furthermore RelSim (\ref{eq:relsim}) depends only on the stored prototypes $W$ and the new unlabelled data point $x$ and no additional storage is needed.

\textbf{Bayes:} The Bayes classifier provides class probabilities for each class provided the data distribution is known. The reject option uses the certainty measure
    \begin{equation}
        r_{\text{Bayes}}(x) = \max_{1\leq j \leq Z} p(j|x) \in (0,1]
    \end{equation}
is optimal regarding the error-reject trade-off \cite{chow1970optimum}.
%We use it as ground truth for artificial data with known underlying distribution.
In general, the class probabilities are unknown, such that this optimum Bayes reject option can serve as Gold standard for artificially designed settings with a known ground truth, only.

\section{Evaluation of Reject Options using Reject Curves} \label{sec:eval}
ARCs \cite{nadeem2009accuracy} are the state of the art for comparing classifiers with a reject option and show the accuracy of a classifier as function of either its acceptance or rejection rate. On the $x$-axis, ARCs show acceptance rates calculated as $|X_{\theta}|/|X|$\footnote{$X$: set of all data; $X_\theta$: set of data with certainty values greater or equal to $\theta$; \newline$|\cdot|$ denotes the cardinality of a set. },
given an applied threshold $\theta$, while on the $y$-axis, the corresponding accuracy calculated on $X_{\theta}$ is shown. Similarly, the $x$-axis can show rejection rates $1-|X_{\theta}|/|X|$.

ARCs can be easily calculated for binary and multi-class classification scenarios as long as a reject option can be defined for the classifier in question.
Formally, the ARC for a given binary data set $X$ is defined as
\begin{equation}
    ARC(\theta): [0,1] \rightarrow [0,1], \frac{|X_{\theta}|}{|X|} \mapsto \frac{\mathit{TP_\theta} + \mathit{TN_\theta}}{|X_{\theta}|}
\end{equation}
with $\theta\in\mathbb{R}$, and the true positives ($\mathit{TP_\theta}$) and the true negatives ($\mathit{TN_\theta}$) in $X_\theta$.

While for many classification tasks, in particular for balanced data sets,  the accuracy and hence the ARC are suitable techniques, there are scenarios where other evaluation metrics of the classification performance are preferred.
For instance, in highly imbalanced scenarios the accuracy of a classifier may be high simply due to---in the worst case---constantly predicting the majority class while the minority class is always misclassified.
In such scenarios measures like the $F1$-score, or precision and recall \cite{van1974foundation} avoid misjudging the performance of a classifier on imbalanced data sets.
In \cite{pillai2011classification} a reject curve is proposed with the $F1$-score instead of the accuracy for multi-label settings.
Similarly, we introduce the precision reject curve (PRC) and recall reject curve (RRC) as follows,

\begin{equation}
    PRC(\theta): [0,1] \rightarrow [0,1], \frac{|X_{\theta}|}{|X|} \mapsto \frac{\mathit{TP_\theta}}{\mathit{TP_\theta} + \mathit{FP_\theta}},
   \label{eq:prc}
\end{equation}

\begin{equation}
   RRC(\theta): [0,1] \rightarrow [0,1], \frac{|X_{\theta}|}{|X|} \mapsto \frac{\mathit{TP_\theta}}{\mathit{TP_\theta} + \mathit{FN_\theta}}.
  \label{eq:rrc}
\end{equation}
where $\mathit{FP_\theta}$ and $\mathit{FN_\theta}$ are the false positives and the false negatives in $X_\theta$.
Here we show the application of PRCs and RRCs for binary classification only. Similarly to ARCs for multi-class classification (e.\,g.,~\cite{fischer2015efficient}), both approaches can be extended to multi-class settings, since precision and recall generalize as well \cite{manning2009}.

\section{Experiments}\label{sec:experiments}

We demonstrate the usefulness of the PRC and the RRC for the three types of data sets, artificial Gaussian data with known class probabilities, two benchmark data sets, and the Adrenal data set from a real-world medical application.
% The class probabilities are known for the artificial data and we calculate the ground truth for reject curves using a Bayesian classifier.

\textit{Gaussian Clusters:}
The data set contains two artificially-generated, overlapping 2D Gaussian classes, overlaid with uniform noise. Samples are equally distributed over classes.
Parameters used were means $\mu_x = (-4, 4.5)$ and $\mu_y = (4, 0.5)$, and standard deviations $\sigma_x = (5.2, 7.1)$ and $\sigma_y = (2.5, 2.1)$.

\textit{Tecator data set:} The goal for this data set is to predict the fat content (\textit{high}~vs.~\textit{low}) of a meat sample from its near-infrared absorbance spectrum. Samples are non-equally distributed over classes with \SI{36.0}{\percent} vs. \SI{64.0}{\percent}.

\textit{Haberman Data Set:}
The data set contains \num{306} instances from two classes (\SI{26.5}{\percent} vs. \SI{73.5}{\percent}) indicating the survival of \num{5} years and more after breast cancer surgery \cite{uci}.
Three attributes are present: \textit{age}, \textit{year of operation}, and the \textit{number of positive auxiliary nodes} detected.

\textit{Adrenal:}
The adrenal tumours data set \cite{arlt2011urine} comprises \num{147} samples composed of \num{32} steroid marker values as features. The steroid marker values are measured from urine samples using gas chromatography/mass spectrometry.
The data comprises two unbalanced classes, namely, patients with benign adrenocortical adenoma (\num{102} or \SI{68.4}{\percent} samples) and patients with malignant carcinoma (\num{45} or \SI{30.6}{\percent} samples). For medical details we refer to \cite{arlt2011urine,BiehlSSSTHSSA12}.

We use a \num{10}-fold repeated cross-validation with ten repeats for our experiments and evaluate models obtained by RSLVQ, GMLVQ, and LGMLVQ with one prototype per class.
Since RSLVQ provides probability estimates, we use the certainty measure Conf (\ref{eq:conf}) for rejection.
In turn, GMLVQ and LGMLVQ lend itself to the certainty measure RelSim (\ref{eq:relsim}).
%In Figures~\ref{fig:gauss} to \ref{fig:adrenal}, we show ARC, PRC, and RRC averaged over \num{100} runs per data set type and classifier.

\begin{figure}[hb]
    \centering
    \includegraphics[width=\textwidth]{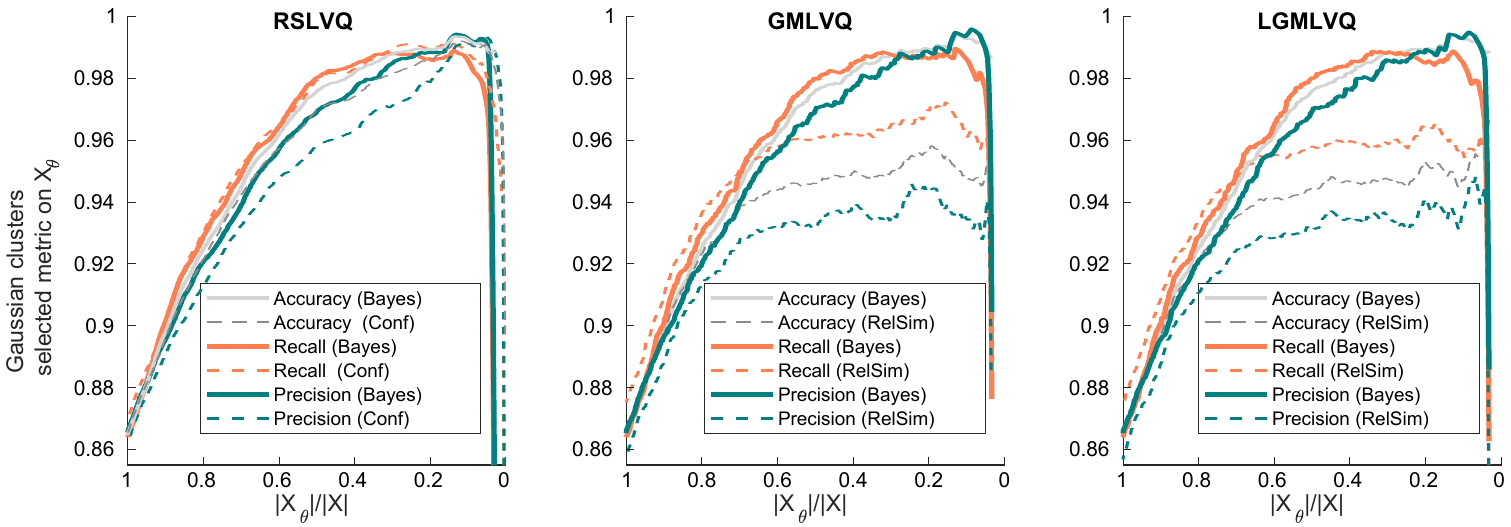}
    \caption{The averaged reject curves for the different models of the artificial Gaussian data are shown (mean over models in different runs). The solid lines represent the optimal classification performance of a Bayesian classifier. The PRCs and RRCs based on RelSim or Conf perform similar to the optimal ARCs \cite{FischerHW14,fischer2016optimal} for the important regime of at least $80\,\%$ accepted data points.}
    \label{fig:gauss}
\end{figure}

% In  Fig.~\ref{fig:gauss}, we show ARC, RRC, and PRC of the Bayesian classifier as well as for the trained prototype models for the Gaussian data set.
For the Gaussian data (Fig.~\ref{fig:gauss}), the ARC, PRC, and RRC of the RSLVQ model resemble the respective optimal choice obtained from Bayesian classifier closely%. Additionally, the RRCs and PRCs closely follow the baseline shape of the ARCs
, for nearly all rejection rates. This is due to little noise and little overlap in the simulated data. For the GMLVQ and the LGMLVQ the shapes of the ARCs, PRCs, and RRCs based on RelSim or Conf are similar to the respective Bayesian baseline results up to a rejection rate of $0.2=(1-|X_{\theta}|/|X|)$, i.\,e. an acceptance rate of  $|X_{\theta}|/|X|=0.8$. Performance decreases for lower acceptance rate, which, however, may be less relevant in applications.
%For lower  acceptance rates ($|X_{\theta}|/|X|<0.8$), all three reject options do not lead to substantial additional performance improvements. However, acceptance rates below $0.8$ may be irrelevant for applications.
In sum, our proposed reject curves mirror the optimal performance of the Bayesian classifier closely for acceptance rates that can be considered relevant for practical applications.

\begin{figure}[ht]
    \centering
    \includegraphics[width=\textwidth]{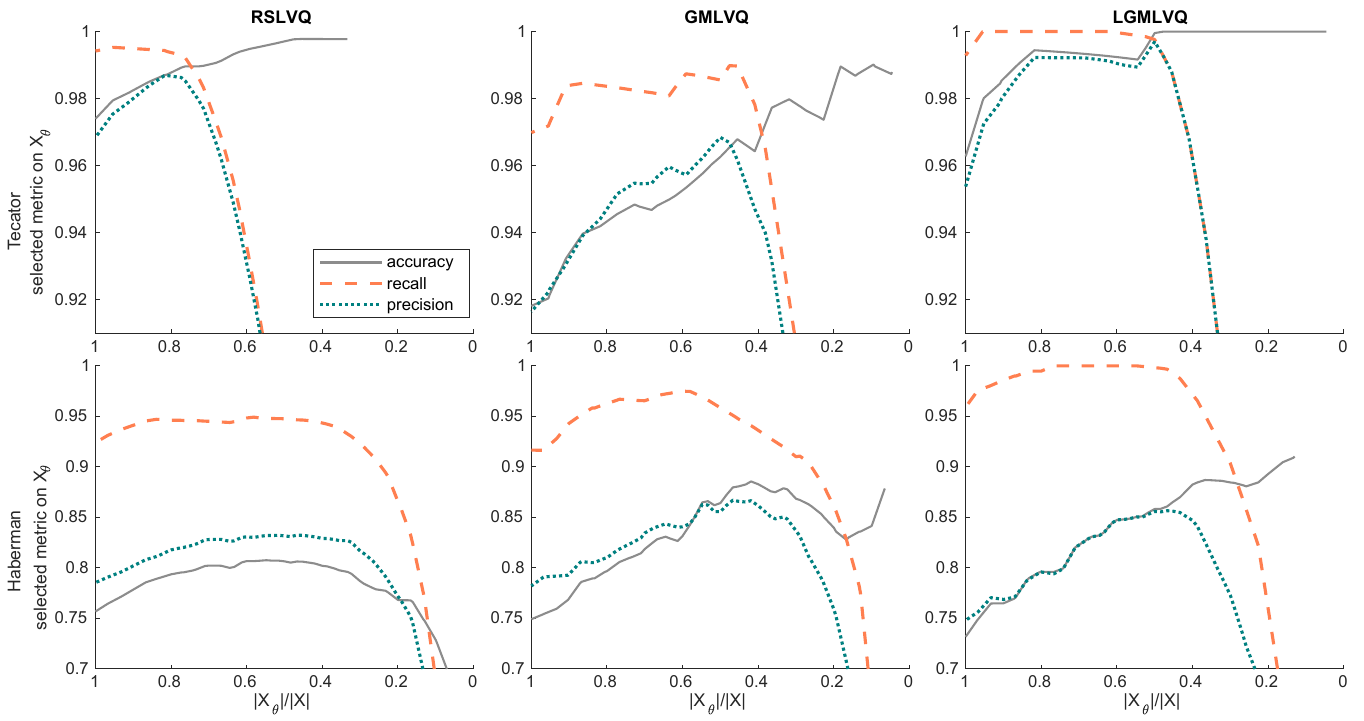}
    \caption{The image shows results for different LVQ models on benchmark data.
    The ARCs \cite{FischerHW14,fischer2016optimal} serve as comparison.
    The PRCs and the RRCs based on RelSim or Conf perform differently for the given set-ups.
    This reveals interesting insights for the user in order to chose a suited reject threshold for the application scenario at hand.
    }
    \label{fig:benchmark}
\end{figure}

%Fig.~\ref{fig:benchmark} shows reject curves for the
For the benchmark data (Fig.~\ref{fig:benchmark}), RRCs and PRCs are compared to ARCs \cite{FischerHW14,fischer2016optimal} .
Precision is in the same range as accuracy in case of no rejection and for high and medium rejection rates ($|X_{\theta}|/|X|>0.3$ for Tecator and $|X_{\theta}|/|X|>0.1$ for Haberman).
Recall has higher values in case of no rejection and behaves similarly to precision for rejection rates greater than zero. For lower rejection thresholds~$\theta$ the shapes of the RRCs and the PRCs are monotonically decreasing, with the effect being most prominent for the Tecator data. This behavior can be expected as precision and recall focus on one of the classes instead of both (in binary classification).
Here we see that ARC is not a suitable tool to meaningfully evaluate model performance across the full range of $\theta$, as the accuracy overestimates the performance as shown by applying the PRC and RRC.
% Instead while PRC and RRC allow to evaluate model performance for a specific threshold $\theta$ with respect to the more appropriate measures recall and precision.

\begin{figure}[ht]
    \centering
    \includegraphics[width=0.69\textwidth]{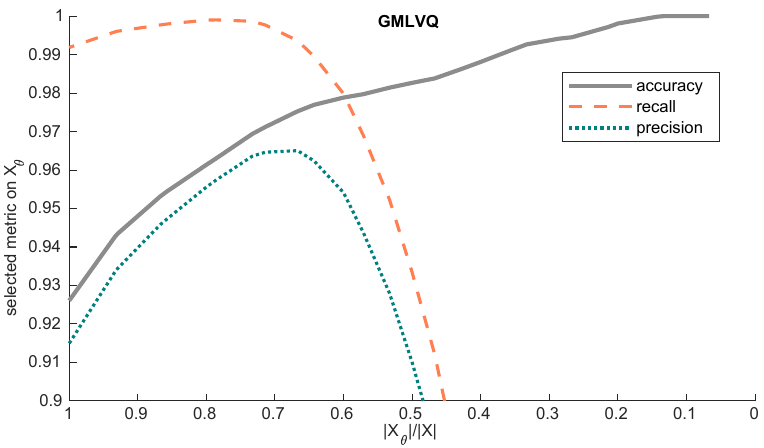}
    \caption{The averaged curves of the ARC \cite{fischer2016optimal} and PRC perform similar in the important regime of at least $80\,\%$ accepted data points for the Adrenal data while the RRC has a different shape. The RelSim is used as measure.}
    \label{fig:adrenal}
\end{figure}

Reject curves are relevant for safety-critical scenarios, e.\,g.\ in the medical domain.
Hence, our last experiment demonstrates the application of the proposed reject curves on a real-world medical data set (Fig.~\ref{fig:adrenal}).
% We observe similar results as for the benchmark data sets---
Similar to the benchmark data set, for higher thresholds $\theta$, precision and recall decline while accuracy increases, thus over-estimating performance for low acceptance rates. The experiment shows again that the ARC is not suitable for applications with imbalanced data, while PRC and RRC allow to select meaningful rejection thresholds.

Note that in all experiments, precision and recall develop in a similar fashion. Accordingly, the F1-score will behave in a qualitatively similar fashion as it is the harmonic mean of both quantities. We therefore do not show F1 reject curves here as they will not add insights beyond the PRC and RRC.

\section{Conclusion}\label{sec:conclusion}

In this paper we introduced the precision reject curve (PRC) and the recall reject curve (RRC) to evaluate reject options for classification tasks in which precision and recall are the preferred evaluation metrics over the accuracy (e.\,g., for imbalanced data sets).
%We compare our proposed approach against the state-of-the-art evaluation using accuracy reject curves (ARC).
To demonstrate the suitability of the proposed PRC and RRC, we applied both methods, first, to an artificial data set where we obtained a performance close to ground-truth solutions obtained from Bayesian classifiers. Further, we applied our approach to two popular classification benchmarks, where we showed that our proposed approach allows additional insights into the performance of a classifier on imbalanced data, which could not be obtained from state-of-the-art evaluation using ARCs. Last, we applied the PRC and RRC to one real-world data set from the medical domain where trust in the classification result is particularly important. The latter experiment demonstrates the applicability of our approach as well as its usefulness in a real-world application domain with imbalanced data.
Our results show that when evaluating reject options for classification of imbalanced data, the ARC may be misleading, while PRC and the RRC provide meaningful comparisons.
Future work may extend the proposed approach to multi-class classification and further evaluation metrics. % other evaluation metrics, e.\,g.\ true positive and false positive rates.

\bibliographystyle{splncs04}
\bibliography{reject_curves}

\end{document}